\pdfoutput=1 

\documentclass[letterpaper, 10 pt, conference]{ieeeconf}  

\IEEEoverridecommandlockouts                              

\usepackage{bm}
\usepackage{cite}
\usepackage{flushend}
\include{preamble}

\title{
\LARGE \bf
Five-fingered Hand with Wide Range of Thumb\\Using Combination of Machined Springs and Variable Stiffness Joints
}

\author{Shogo Makino, Kento Kawaharazuka, Ayaka Fujii, Masaya Kawamura, Tasuku Makabe,\\Moritaka Onitsuka, Yuki Asano, Kei Okada, Koji Kawasaki, Masayuki Inaba$^{1}$%
\thanks{$^{1}$The authors are with the Department of Mechano-Informatics, The University of Tokyo 7-3-1 Hongo, Bunkyo-ku, 113-8656 Tokyo, Japan
        {\tt\small \{makino, kawaharazuka, a-fujii, kawamura, makabe, onitsuka, asano, k-okada, inaba\}@jsk.t.u-tokyo.ac.jp, koji\_kawasaki@mail.toyota.co.jp}}
}

\begin{document}

\setlength{\pdfpageheight}{11in}
\setlength{\pdfpagewidth}{8.5in}

\maketitle
\thispagestyle{empty}
\pagestyle{empty}

\begin{abstract}
Human hands can not only grasp objects of various shape and size and manipulate them in hands but also exert such a large gripping force that they can support the body in the situations such as dangling a bar and climbing a ladder. On the other hand, it is difficult for most robot hands to manage both.
Therefore in this paper we developed the hand which can grasp various objects and exert large gripping force. To develop such hand, we focused on the thumb CM joint with wide range of motion and the MP joints of four fingers with the DOF of abduction and adduction.
Based on the hand with large gripping force and flexibility using machined spring, we applied above mentioned joint mechanism to the hand.
The thumb CM joint has wide range of motion because of the combination of three machined springs and MP joints of four fingers have variable rigidity mechanism instead of driving each joint independently in order to move joint in limited space and by limited actuators.
Using the developed hand, we achieved the grasping of various objects, supporting a large load and several motions with an arm.
\end{abstract}

\section{INTRODUCTION}

Human hands can grasp objects of various shape and size and manipulate them in the hand as it wished.
And what is more, its gripping force is so large that it can support their body in the situations such as dangling a bar and climbing a ladder.
Inspired by human hands, many human-mimetic robot hands have been developed.
As dexterous hands, the hand which have as large DOFs as human hands and which can control those DOFs dependently have been developed\cite{industrialrobot:kochan:shadow_hand, icra2011:dlr_hand_arm, icra2014:roboray_hand:kim}.
Some hands have flexible structure of soft material or compliant joints and can move along various objects\cite{ijrr2016:Deimel:compliant_hand, icra2017:compliant_hand:Wiste}.
As another example, there is a hand imitating human in the point of not only the whole shape and arrangement of joints but also joint structure consisting of the parts in the shape of bones and ligaments made of lopes\cite{icra2016:ligament_hand:xu}.
As high-power hand, the hand which has grasping-force-magnification mechanism using eccentric cam\cite{takaki:high_power_hand} has been developed.
Authors developed the hand which has robust and flexible finger joints using machined springs and succeeded in making the life-sized humanoid ``Kengoro''\cite{humanoids2016:asano:kengoro}, whose weight is 56.2 kg, dangle a bar\cite{HandDesign:Makino:IROS2017}.
However, little human-mimetic hands have both dexterity and high-power.
Dexterous and high-power hands are demanded as hands of humanoid robots as an example.
Humanoid robots are expected to perform a variety of tasks in the environment we live in and cannot enter and by using the tool we use because of the physical characteristics that their bodies are similar to human beings.
In DARPA Robotics Challenge held in 2015, humanoid robots were needed to perform tasks to use hands such as driving and getting off a car, drilling a hole and turning a valve.
To perform such various tasks, the hands should have enough dexterity to use various tools and enough power to support and pull the body.

Therefore, in this paper we developed the hand which can grasp various objects and support a large load.
To develop such hand, we focused on the movability of the CM joint of thumb and the MP joints of four fingers.
We developed the thumb CM joint with wide range motion with the combination of three machined springs and we applied the thumb CM joint and variable rigidity mechanism using agonistic tendon drive in the MP joints of four fingers to the robust and flexible hand using machined springs.
In this way, we developed the hand with both of dexterity and high-power.

\begin{figure}[t]
 \begin{center}
  \includegraphics[width=0.85\columnwidth]{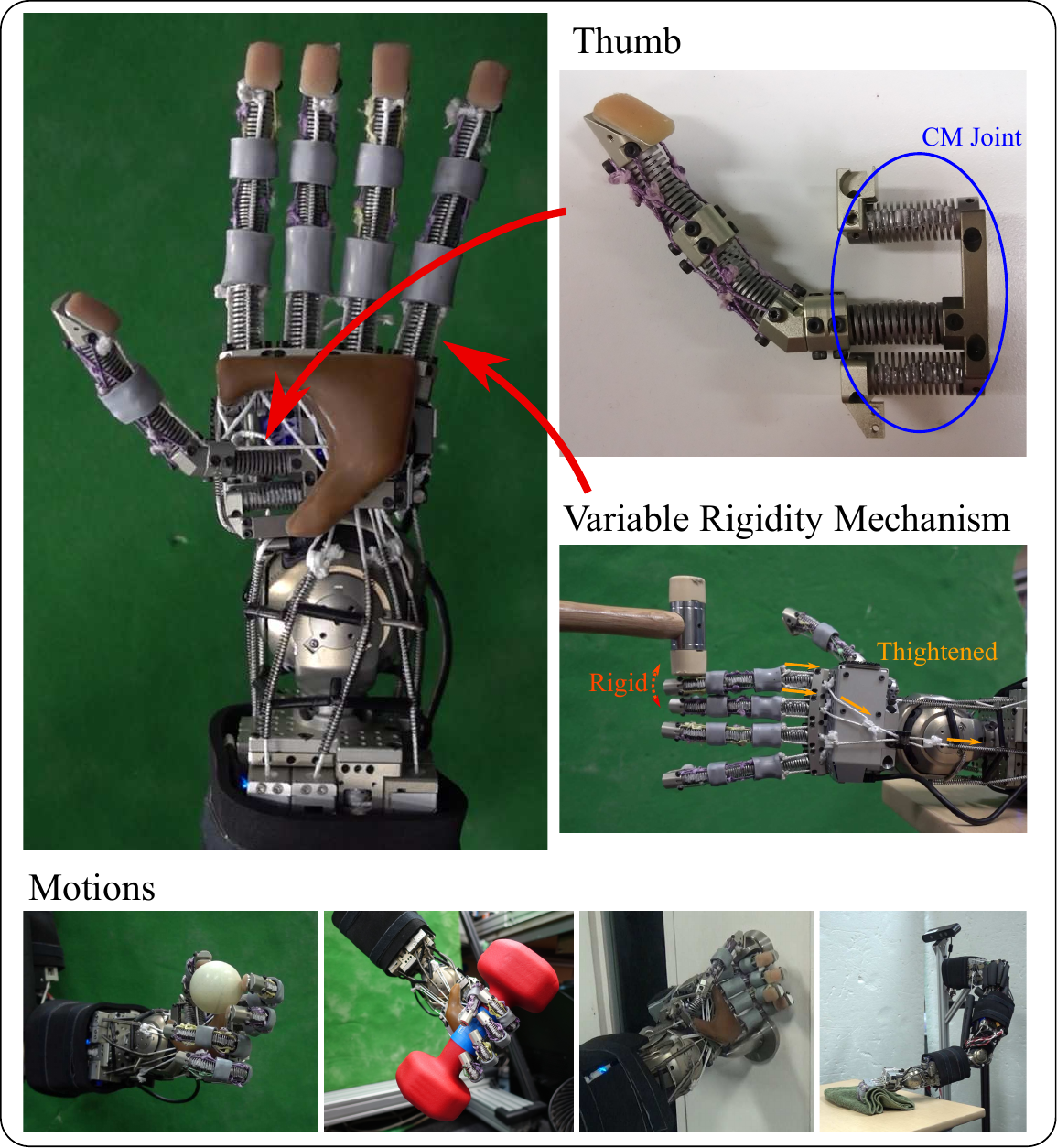}
  \caption{Developed hand in this study.}
  \label{figure:cover_picture}
 \vspace{-3.0ex}
 \end{center}
\end{figure}

In section I, we explained the background of the study.
In section II, we will explain the structure of human hands and the mechanism to which the hand is applied.
In section III, we will explain the design and fabrication of the hand in detail.
In section IV, we will explain the experiments with the developed hand with an arm.
In section V, we will state the conclusion of this study and future works.

\section{STRUCTURE OF HUMAN HANDS AND APPLICATION TO ROBOT HANDS}
In this section, we will discuss the structure of human hand and the method to apply the structure to robot hands.
\subsection{The Structure of Human Hand}
\begin{figure}[t]
 \begin{center}
  \includegraphics[width=0.55\columnwidth]{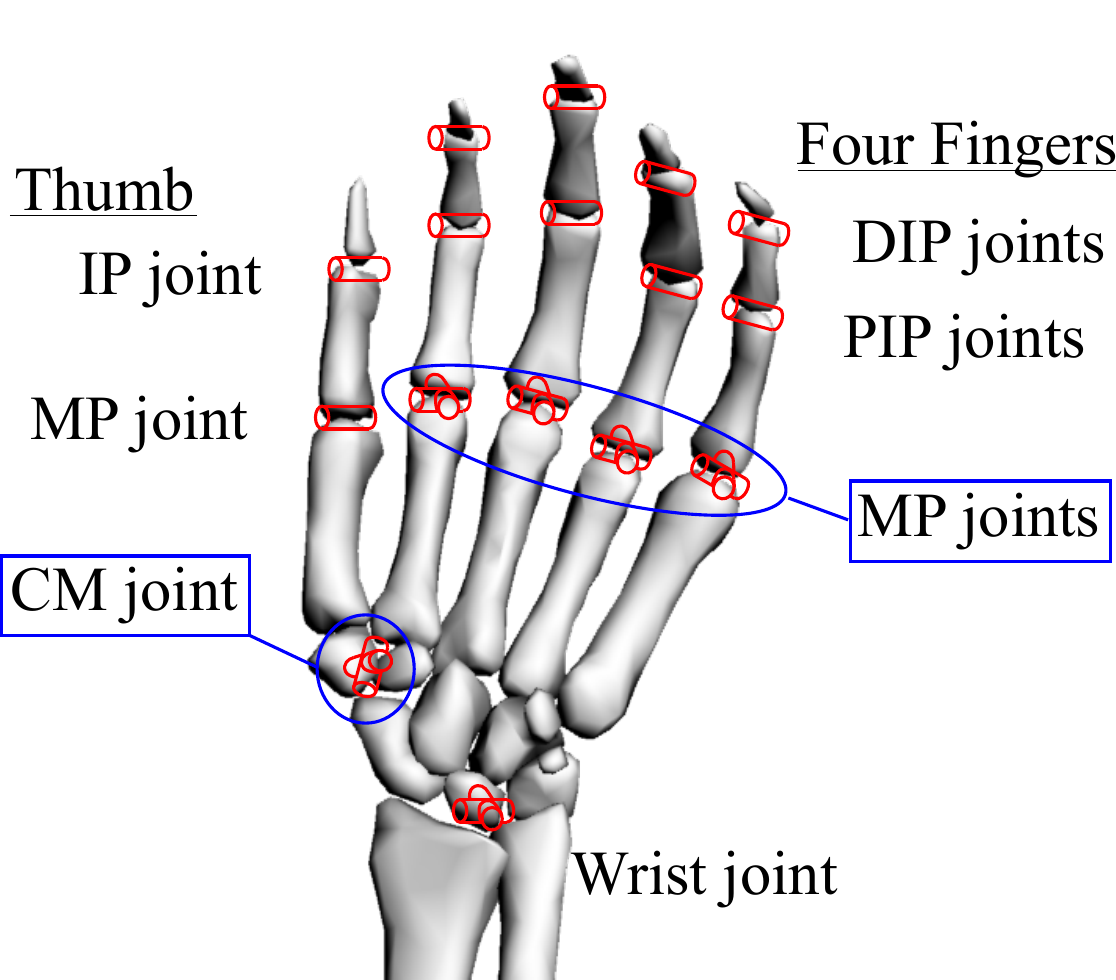}
  \caption{Structure of human hand.}
  \label{figure:human_hand}
 \vspace{-3.0ex}
 \end{center}
\end{figure}

Needless to say, the DOFs of flexion of fingers are necessary for grasping objects with fingers.
However, other DOFs in hands play an important role upon manipulating various objects dexterously.
The structure of human hand is shown in \figref{human_hand}

First, we will focus on the thumb.
The thumb CM joint has two DOFs: opposition and abduction-adduction.
It should be noted that the DOF of opposition has such wide range of motion that the hand can be made a plane with open hands and the shape that the fingertips of thumb and little finger are touched. The combination of the two DOFs enables the fingertip of thumb to take an every position around the palm and fingers. Human beings change the angle of this joint depending on the situations, for example, completely extended when wiping table, about 90 degree when grasping a cylinder and more than 90 degree when grasping thin cloth or rotating small ball in hand.

Next, we will focus on the MP joints of four fingers, which are the boundary joints between palm and fingers.
This joint also have two DOFs: flexion-extension and abduction-adduction.
The DOFs of abduction and adduction are moved by palmar interossei muscle and dorsal interossei muscle.
These two kind of muscles can adjust the position of fingertips depending on the shape of objects and how to hold objects.
In addition, they enables fingers to transmit force with the side of finger such as operating a lever in a vehicle and turning a key by increasing the rigidity of MP joints.

\subsection{Application to Robot Hands}
Next, we will discuss how we apply characteristics of human hand mentioned above to human mimetic robot hands.
We aimed at developing a dexterous and high-power hand by adding dexterity to high-power hand using machined spring.
But there are some problems for adding.
Concerning thumb CM joint, it is difficult to make a joint with wide range of motion in the limited space or one joint can affect to another joint in the case of the joint with two DOFs including a DOF with wide range of motion because the spring requires a long coil to secure enough range of motion.
Concerning MP joints, it is difficult to arrange actuators to control muscles independently because there are eight interossei muscles to move four fingers in quite a narrow space.

Therefore, we made a novel thumb CM joint by the combination of three machined springs and realized the joint with DOFs including a joint with wide range of motion.
And we applied a variable rigidity mechanism of agonistic tendon drive and placed one actuator to move this mechanism toward both of the fit of finger when grasping and the transmission of force with the side of fingers.

\section{DESIGN AND FABRICATION OF THE HAND}
\subsection{Overview}
\begin{figure}[t]
 \begin{center}
  \includegraphics[width=0.98\columnwidth]{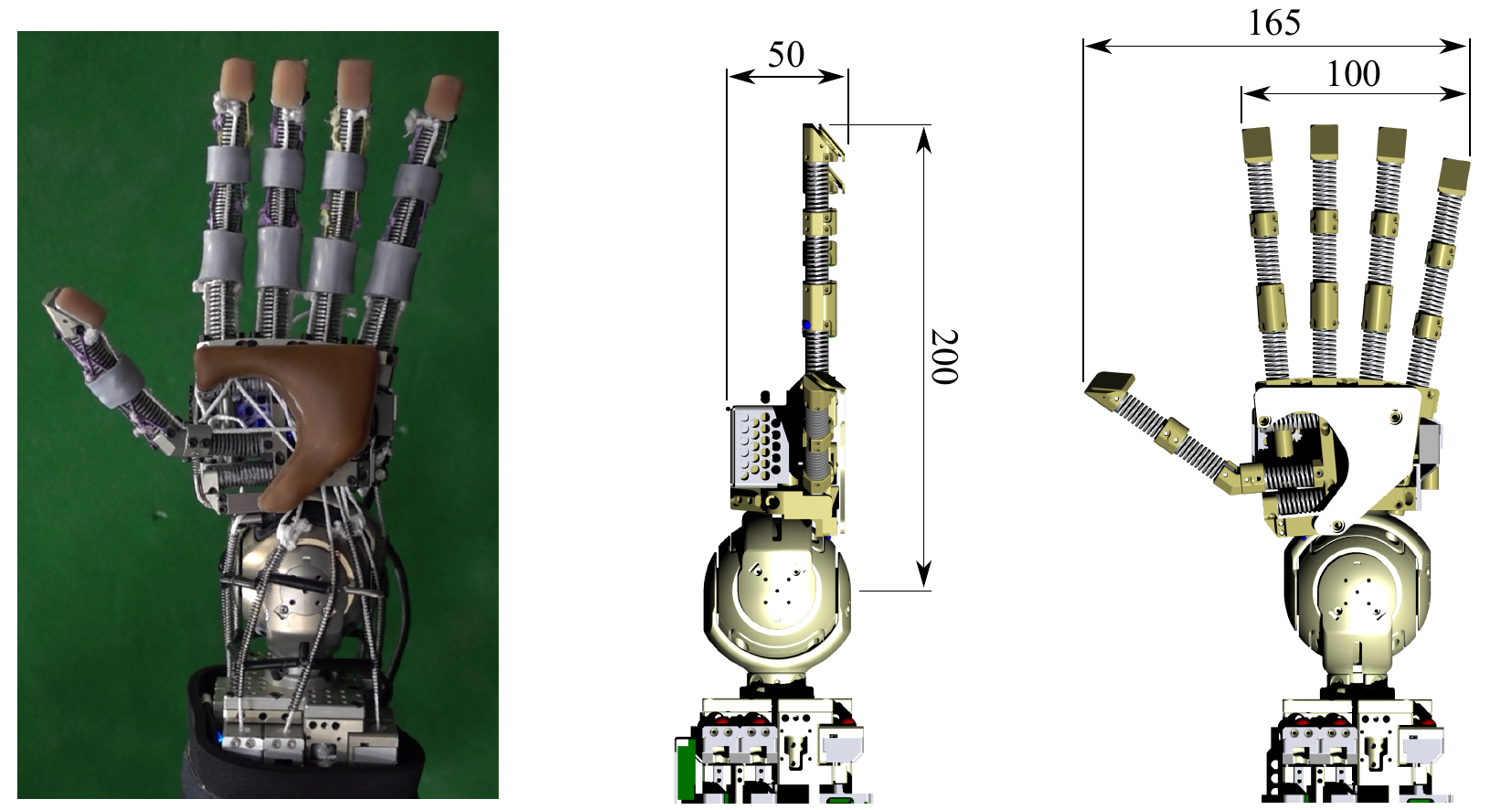}
  \caption{The size of newly developed hand.}
  \label{figure:hand_overview}
  \vspace{-3.0ex}
 \end{center}
\end{figure}

The whole view and size of the hand we developed in this study are shown in \figref{hand_overview}.
It is a five-fingered hand imitating human hands.
It is connected to forearm through the wrist joint and it is moved by the muscles placed in forearm.

\subsection{Finger Joints Using Machined Spring}
\begin{figure}[t]
 \begin{center}
  \includegraphics[width=0.98\columnwidth]{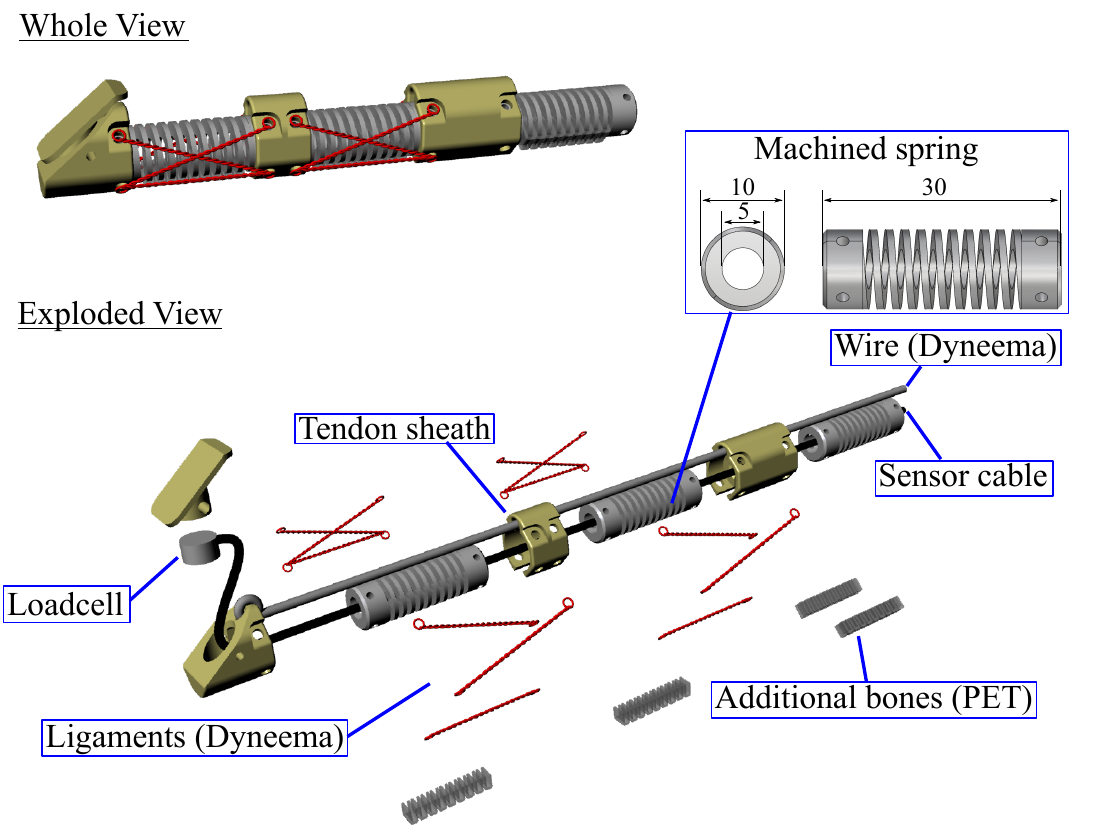}
  \caption{Detail of fingers.}
  \label{figure:finger}
 \end{center}
\end{figure}

The fingers of this hand is shown in \figref{finger}.
It has finger joints made of machined springs as with the hand of ``Kengoro''.
The machined spring has flexibility and toughness because it is a kind of springs made of metal.
In addition, it has an advantage that it can be connected firmly to other parts because it can be designed integrally with attachments.
For these reason, it is suitable for finger joints of the hand, which is needed to support a large load and withstand impacts.
Therefore, we made the fingers consists of three machined springs and the parts to connect springs imitating tendon sheath.
In addition, the machined spring has another merit of the independence in design between torsional and bending stiffness.
We took account of only the DOF of bending because it can be designed with enough large torsional stiffness compared to the bending stiffness.
The bending stiffness of finger joints is shown in \tabref{spring_stiffness}.

\begin{table}[t]
  \begin{center}
    \caption{Spring constants of machined springs in each joint[deg/Nm].}
    \label{table:spring_stiffness}
    \begin{minipage}{0.6\linewidth}
      \begin{tabular}{|c|c|}
        \multicolumn{2}{l}{Thumb}\\
        \hline
        & Stiffness\\
        \hline
        CM Joint & 903\\
        & (Parallel two springs$^{*1}$)\\
        & 443\\
        & (The third spring$^{*1}$)\\
        \hline
        MP Joint & 443\\
        \hline
        IP Joint & 443\\
        \hline
        \multicolumn{2}{l}{$^{*1}$ Mentioned in section III-C}\\
      \end{tabular}
    \end{minipage}
    \begin{minipage}{0.35\linewidth}
      \begin{tabular}{|c|c|}
        \multicolumn{2}{l}{Four fingers}\\
        \hline
        & Stiffness\\
        \hline
        MP Joint & 664\\
        \hline
        PIP Joint & 863\\
        \hline
        DIP Joint & 443\\
        \hline
      \end{tabular}
    \end{minipage}
  \end{center}
\end{table}

\begin{figure}[t]
 \begin{center}
  \includegraphics[width=0.85\columnwidth]{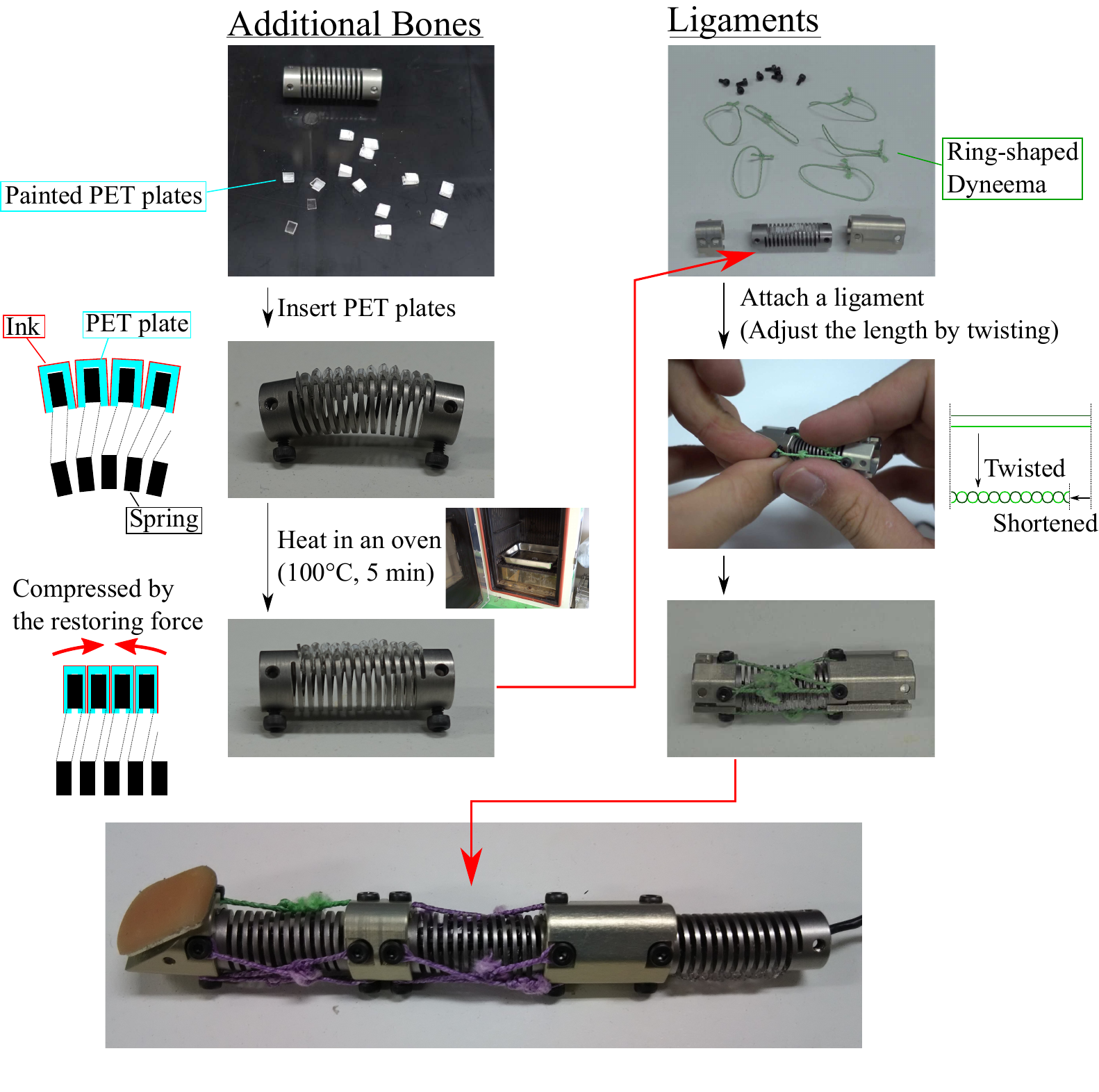}
  \caption{Fabrication of finger.}
  \label{figure:finger_fabrication}
  \vspace{-3.0ex}
 \end{center}
\end{figure}

Next, we will mention additional parts placed around machined springs.
These parts are placed to inhibit extra DOFs of spring joints.
The DOF of bending is separated two kind of DOFs: flexion-extension and abduction-adduction.
Also, spring joints have another DOF: elongation-shrinkage.
For the use in finger joints, these DOFs except flexion-extension have to be restricted.
Therefore, we applied additional bones made of PET to inhibit shrinkage and ligaments made of Dyneema to inhibit elongation.
The combination of them also inhibit the other rotational DOF.
The way of fabrication of fingers is shown in \figref{finger_fabrication}.
The thickness of PET plates is naturally adjusted by heating in a oven.
The length and tension of ligaments is adjusted by twisting wire.

\subsection{Thumb CM Joint with wide Range of Motion}
\begin{figure}[t]
 \begin{center}
  \includegraphics[width=0.85\columnwidth]{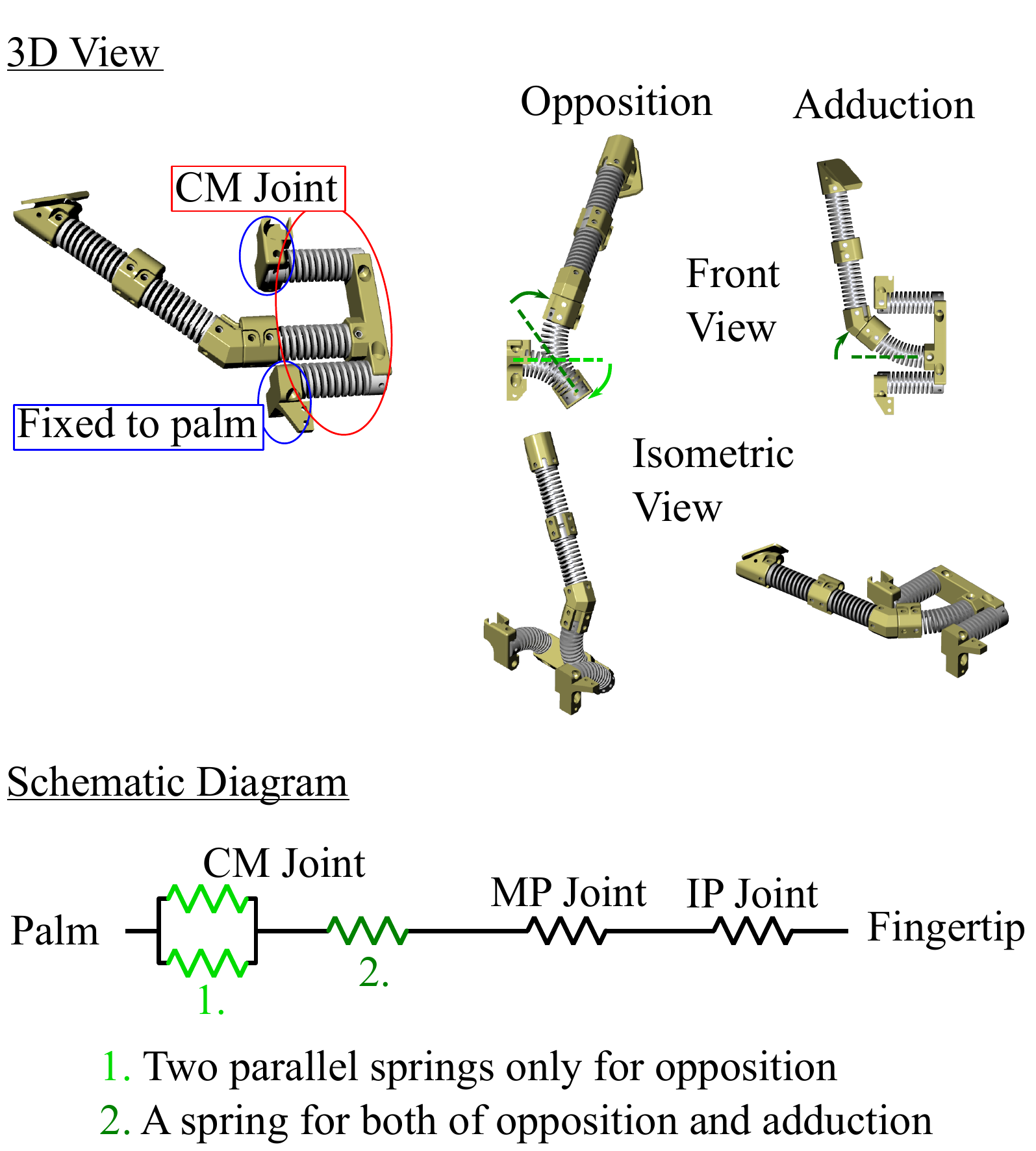}
  \caption{Detail of the thumb.}
  \label{figure:thumb}
 \end{center}
\end{figure}

As mentioned in previous section, the CM joint has two remarkable characteristics.
\begin{itemize}
  \item It has two DOFs: opposition and abduction-adduction.
  \item The DOF of opposition has wide range of motion.
\end{itemize}
To realize these characteristics in the limited space, this joint consists of three machined springs as shown in the top of \figref{thumb}.
The CM joint is in the position of a red circle, and it is connected to the palm in the position of blue circles.
The schematic diagram of spring arrangement is shown in the bottom of \figref{thumb}.
The two of three springs are placed in parallel and this pair and the third spring are connected in series, turned back in the connected point.
The pair of parallel springs are bended when oppositing and the third spring are bended when both oppositing and adducting.
In the other words, when adducting only one spring is bended as the flexion of other general joints, but when oppositing two serial pairs of springs are bended.
In this way the joint with wide range of motion is realized without little affection to other joints.

\subsection{Variable Rigidity Mechanism by Agonism Tendon Drive}
\begin{figure}[t]
 \begin{center}
  \includegraphics[width=0.85\columnwidth]{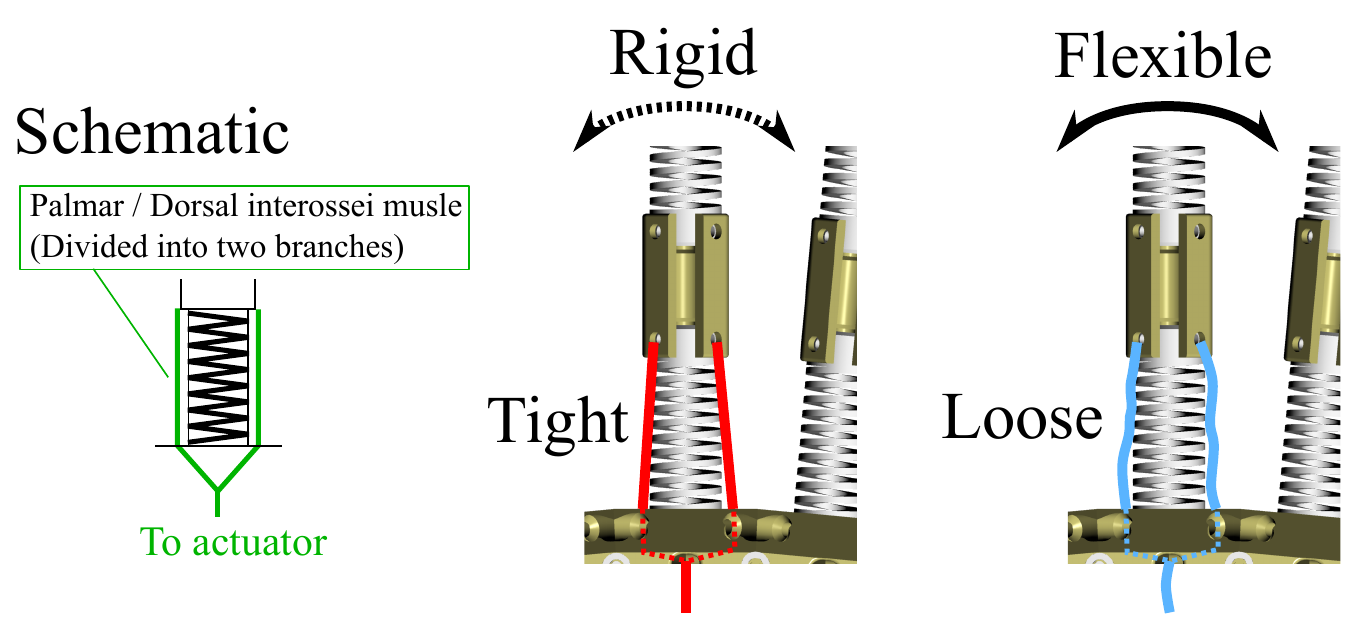}
  \caption{Variable rigidity mechanism.}
  \label{figure:variable_rigidity}
 \end{center}
\end{figure}

The MP joints of four fingers have variable joint stiffness mechanism as shown in \figref{variable_rigidity}.
The wire from an actuator branches to palmar interossei muscle and dorsal interossei muscle.
The rigidity of the rotational DOF of abduction and adduction is changed by tightening these two muscles.
This mechanism is inserted in four fingers, but the only one actuator drives these muscles because the wires of four fingers are connected on the back of the hand.
This mechanism cannot move MP joints actively but it is expect for fingers to bend along objects when the wire is loosen to make the rigidity of MP joints low and to transmit a force with the side of finger when the wire is tighten to make the rigidity high.

\subsection{Other Characteristics}
\subsubsection{Wire Arrangement}
\begin{figure}[t]
 \begin{center}
  \includegraphics[width=0.98\columnwidth]{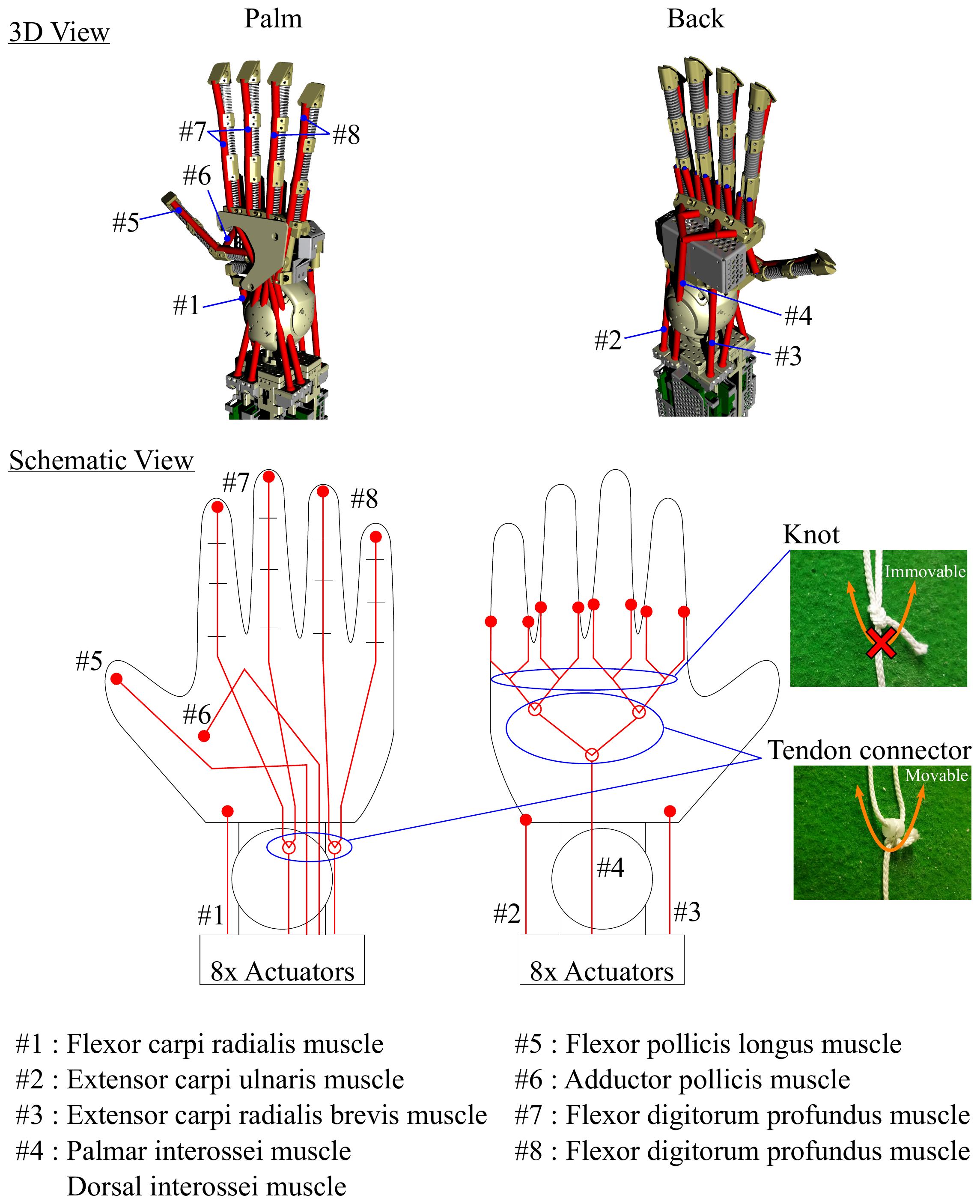}
  \caption{Wire arrangement.}
  \label{figure:wire_schematic}
 \end{center}
\end{figure}

The actuators to move this hand are the miniature bone-muscle modules\cite{ForearmDesign:Kawaharazuka:IROS2017} in forearm.
The motors included in modules are all Maxon EC-16 60W 128:1.
The number of actuators is eight because of the limitation of the space in order not to break the proportion of human forearm.
The three of eight muscles are used to move wrist joints which have two DOFs.
The other five muscles are used to move fingers.
The arrangement of wire in 3D model and schematic diagram is shown in \figref{wire_schematic}.
Although the pair of index finger and middle finger, that of ring finger and little finger and the variable rigidity mechanism mentioned in the previous section are respectively moved by one actuator because of the limitation of the number of actuators, by knots in the branch point of palmar interossei muscle and dorsal interossei muscle and ring-shaped tendon connectors\cite{HandClutteredNarrow:Hasegawa:IROS2017} in the other branch points.
It should be noted that all muscles in this hand are placed in forearm although interossei muscles and the adductor pollicis muscle of human beings are intrinsic muscles because of the limitation of the size of actuators and the hand.

\subsubsection{Sensor Arrangement}
\begin{figure}[t]
 \begin{center}
  \includegraphics[width=0.7\columnwidth]{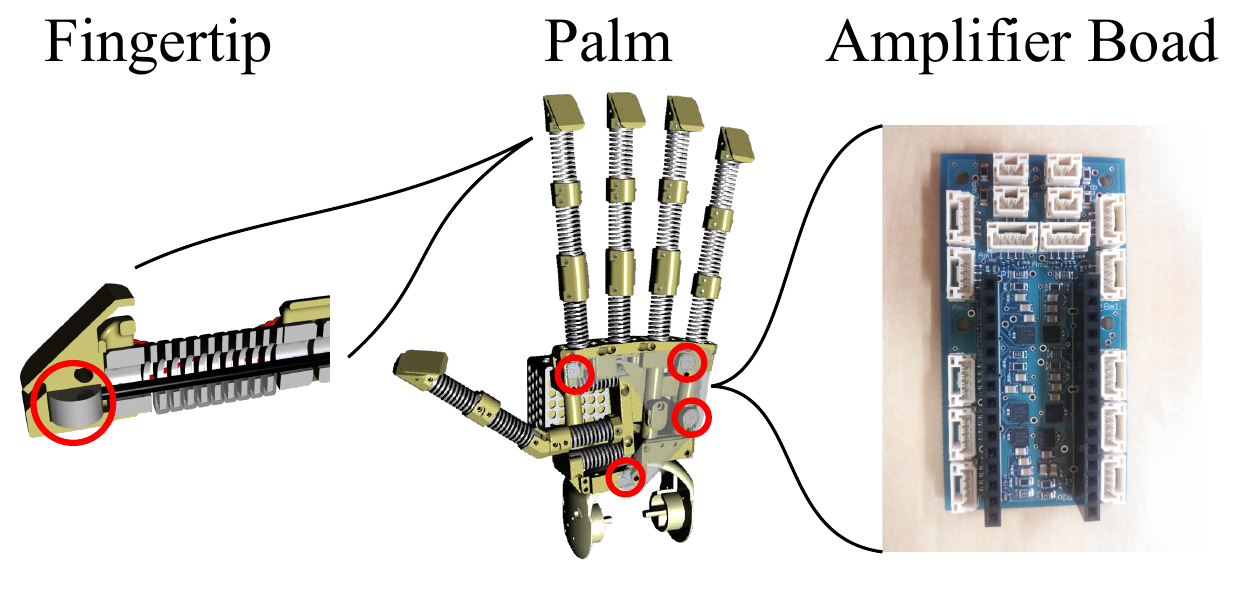}
  \caption{Sensor arrangement.}
  \label{figure:loadcell}
 \end{center}
\end{figure}

Nine loadcells are inserted in this hand.
The position of loadcells is as shown in \figref{loadcell}.
Four loadcells are placed in the position of red circles in the palm.
The others are placed in fingertips.
The cables of loadcells in fingertips pass in coils of springs and all cables are connected to an amplifier board in the back of the hand.
This hand can detect the touch to the object and measure the distribution of external force in the palm by referencing sensor values.

\subsubsection{Exterior}
This hand has exterior made of rubber to secure the softness and frictional force.
An urethane sponge coated by urethane rubber (A60) is pasted on the palm.
Each fingertip plate is coated by urethane rubber (A30).
Concerning finger bones except fingertips, heat shrinkable tube made of silicon are attached in order to attach new exterior easily because they are frequently necessary to be removed in the maintenance.

\section{EXPERIMENTS}
\subsection{Grasping Objects}
\begin{figure*}[t]
  \centering
  \includegraphics[width=0.75\textwidth]{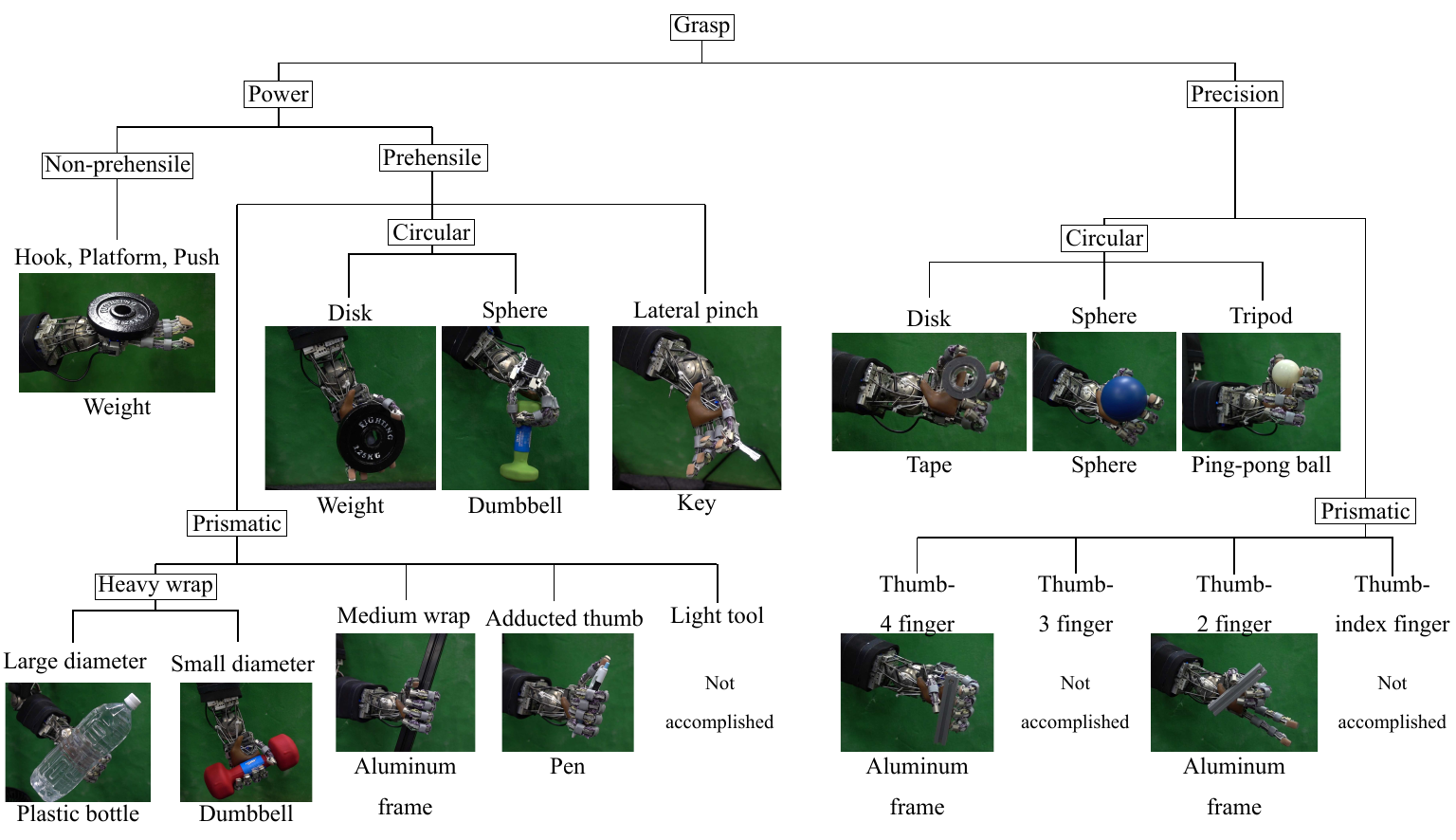}
  \caption{Grasping objects and classification according to \cite{cutkosky:grasp_choice}.}
  \label{figure:grasp_taxonomy}
\end{figure*}

As the basic experiment, we executed the experiment of grasping objects.
In this experiment, we made the array of muscle length in advance depending on the objects heuristically and sent them to the hand.
\figref{grasp_taxonomy} shows the classification of the grasp style depending on the shape and size of objects and the shape of fingers when grasping according to the classification of Cutkosky\cite{cutkosky:grasp_choice}.
``Light tool'' could not be accomplished because it is needed to flex the joint of fingertips but there is only one muscle to flex each finger.
``Thumb-index finger'' and ``Thumb-3 finger'' could not also because it is needed to flex quite differently between two fingers to be flexed by the same actuator for this grasping style.
Although there are some grasp style this hand could not accomplish, this hand could various objects.

\subsection{Supporting a Large Load}
\begin{figure}[t]
  \begin{minipage}{0.5\columnwidth}
    \centering
    \includegraphics[width=0.85\columnwidth]{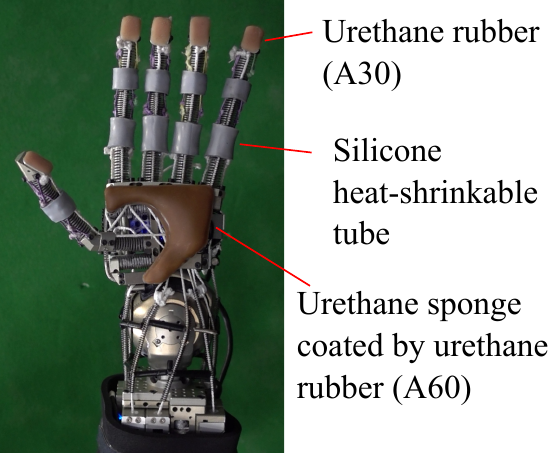}
    \caption{Exterior of this hand.}
    \label{figure:exterior}
  \end{minipage}
  \begin{minipage}{0.45\columnwidth}
    \centering
    \includegraphics[width=0.90\columnwidth]{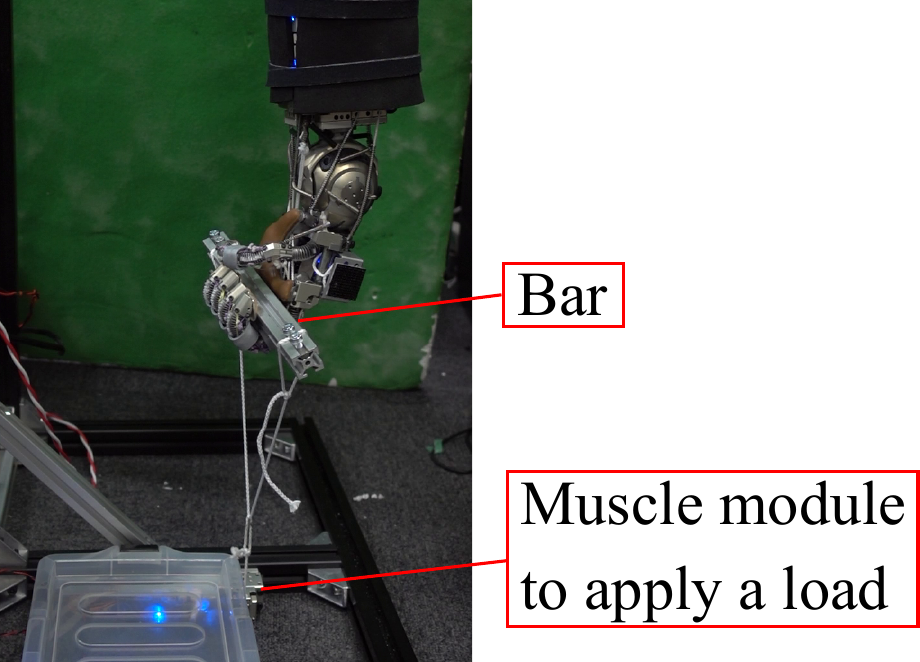}
    \caption{Experiment of supporting a large load.}
    \label{figure:omori}
 \end{minipage}
\end{figure}

\begin{figure}[t]
 \begin{center}
  \includegraphics[width=0.85\columnwidth]{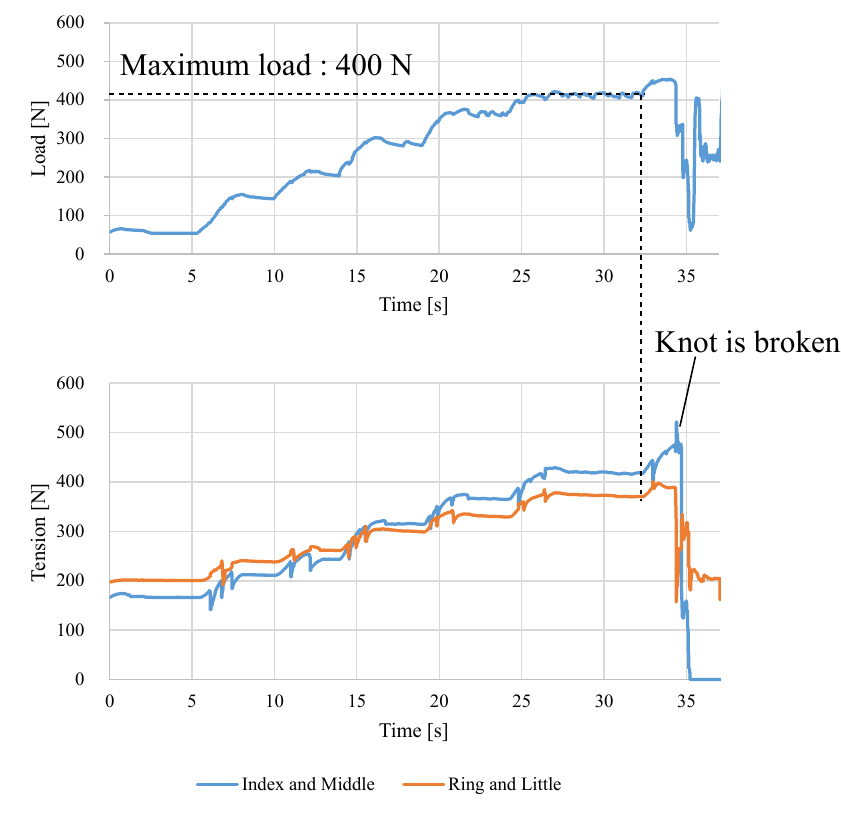}
  \caption{The change of wire tension of fingers.}
  \label{figure:omori_graph}
 \end{center}
\end{figure}

We executed the experiment of supporting a large load to confirm that the developed hand is as powerful as the hand of ``Kengoro'' we developed previously.
First, the hand grasp a bar.
Second, we placed a muscle module\cite{MuscleModule:Asano:IROS2015}, which can exert a force of 50 kgf maximum and can measure the tension of wire, below the hand and we tied the wire from the module to the bar.
Third, the tension of the wire was increased by tightening the wire.
The picture of experiment environment and the change of the tension of the wires for the load and to flex finger is shown in \figref{omori} and \figref{omori_graph} respectively.
The hand could withstand the load of 400 N and after that, a knot was broken and the bar was released.
This hand have enough power to support body but it is expected to support a little bit larger load to improve the connection of wires.

\subsection{Switching a Lever in a Vehicle}
\begin{figure}[t]
 \begin{center}
  \includegraphics[width=0.85\columnwidth]{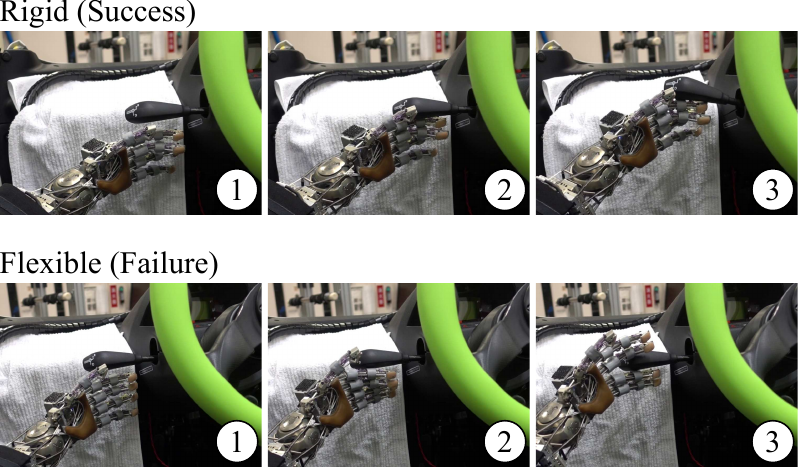}
  \caption{Sequence pictures while switching a lever in a car.}
  \label{figure:winker}
 \end{center}
\end{figure}

As the example of the motion to transmit force with the side of fingers, we executed the experiment of switching a lever in a vehicle.
\figref{winker} shows two set of sequence pictures when the hand was moved from the bottom of the lever to the top of the lever in the condition that the rigidity of MP joints is high and that is low.
The high rigidity enabled the hand to transmit force with the side of finger and to switch a lever.

\subsection{Turning a lock of a door}
\begin{figure}[t]
 \begin{center}
  \includegraphics[width=0.85\columnwidth]{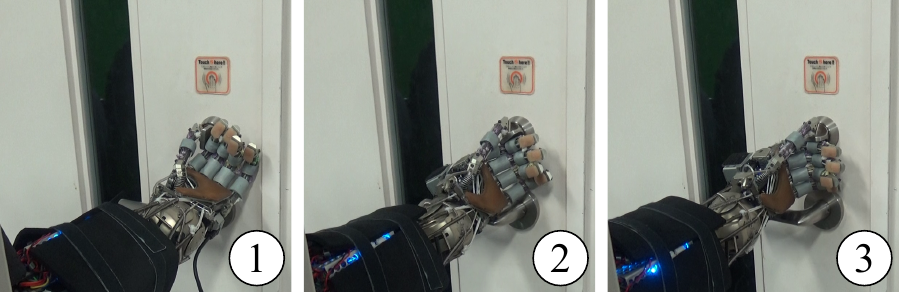}
  \caption{Sequence pictures while turning a lock of a door.}
  \label{figure:thumbturn}
 \end{center}
\end{figure}

As the example of the motion using two DOFs of thumb CM joint and rigid MP joints of four fingers, we executed the motion of turning a lock of a door.
In this experiment, the tension of variable stiffness mechanism is set to 500 N and the hand pinched a knob by the thumb and the side of the index finger.
By rotating forearm, the knob was rotated.
The sequence pictures during this experiment is shown in \figref{thumbturn}

\subsection{Wiping a Table}
\begin{figure}[t]
 \begin{center}
  \includegraphics[width=0.85\columnwidth]{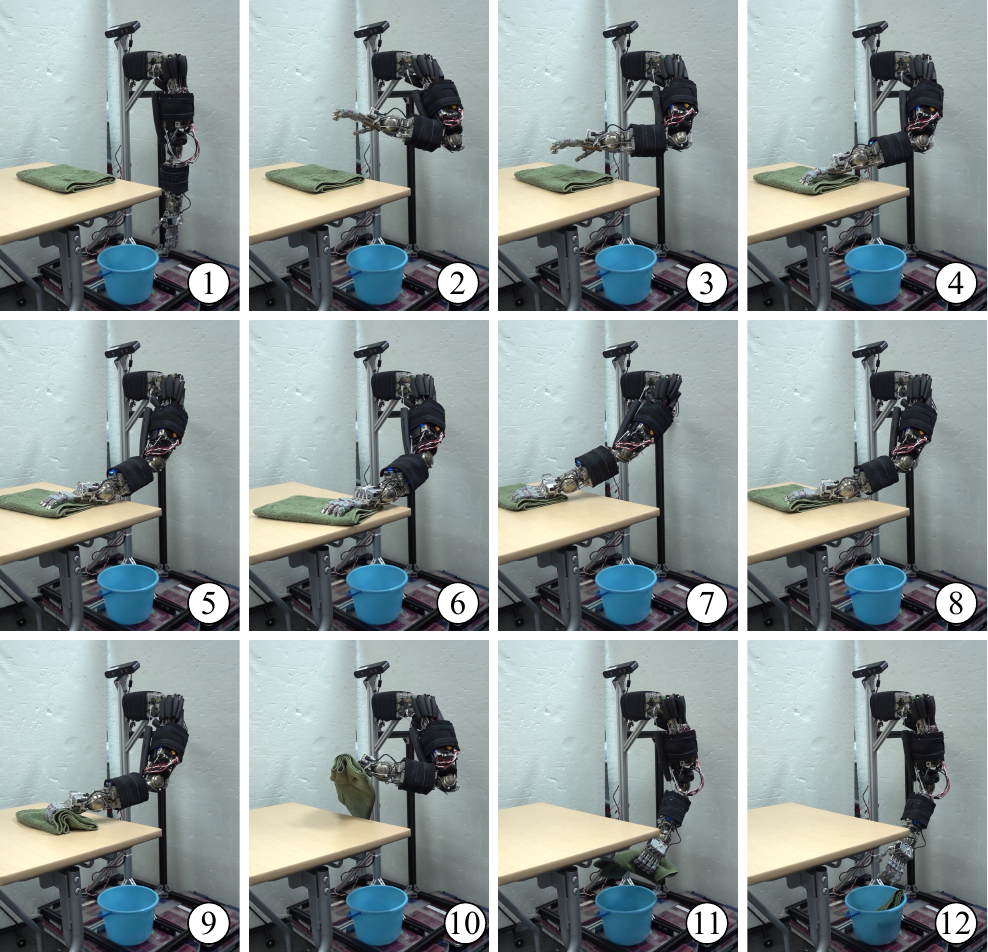}
  \caption{Sequence pictures while wiping a table.}
  \label{figure:wiping}
 \end{center}
\end{figure}

As the example of the motion using the wide range of motion in thumb opposition, we executed the motion of wiping table.
First, the robot, which have the developed hand, found the towel with color filter and moved the hand above the found towel.
Next, the hand was gradually moved down until the sensors in the hand  detect the contact with a threshold value.
After the hand touched the towel, the robot moved the arm to the left and right, while adjusting the angle of wrist joint comparing the force applied in the palm and fingertips.
Finally, the hand grasped the towel, was moved above a bucket put at the predetermined position, drop the towel by opening hand.
Although the hand just moved fingers when grasping, the hand could succeed in grasping without the hand broken because of the compliance and toughness of the hand.
The sequence picture during this motion is shown in \figref{wiping}.

\section{CONCLUSION AND FUTURE WORK}
In this paper, we developed the hand which can exert large gripping force and grasp various objects by applying the hand with flexible and robust finger joint using machined springs to the thumb CM joint with wide range of motion and the variable rigidity mechanism of MP joints of four fingers.
We achieved the motion of switching a lever in a vehicle, turning a lock of a door and wiping a table.
As the future work, we will develop more skillful motion using force sensors in the hand, and tension and length sensors of muscles.
At the same time, we will improve the hardware of the hand such as the consideration of the joint DOFs and the arrangement of the wires, and development of more miniature and high-power actuators.

\addtolength{\textheight}{-20cm}   









\bibliographystyle{IEEEtran}
\bibliography{string, iros2018-makino, common}

\begin{thebibliography}{10}
\providecommand{\url}[1]{#1}
\csname url@rmstyle\endcsname
\providecommand{\newblock}{\relax}
\providecommand{\bibinfo}[2]{#2}
\providecommand\BIBentrySTDinterwordspacing{\spaceskip=0pt\relax}
\providecommand\BIBentryALTinterwordstretchfactor{4}
\providecommand\BIBentryALTinterwordspacing{\spaceskip=\fontdimen2\font plus
\BIBentryALTinterwordstretchfactor\fontdimen3\font minus
  \fontdimen4\font\relax}
\providecommand\BIBforeignlanguage[2]{{%
\expandafter\ifx\csname l@#1\endcsname\relax
\typeout{** WARNING: IEEEtran.bst: No hyphenation pattern has been}%
\typeout{** loaded for the language `#1'. Using the pattern for}%
\typeout{** the default language instead.}%
\else
\language=\csname l@#1\endcsname
\fi
#2}}

\bibitem{industrialrobot:kochan:shadow_hand}
A.~Kochan, ``Shadow delivers first hand,'' \emph{Industrial Robot: An
  International Journal}, vol.~32, no.~1, pp. 15--16, 2005.

\bibitem{icra2011:dlr_hand_arm}
M.~Grebenstein, A.~Albu-Sch{\"a}ffer, T.~Bahls, M.~Chalon, O.~Eiberger,
  W.~Friedl, R.~Gruber, S.~Haddadin, U.~Hagn, R.~Haslinger, H.~H{\"o}ppner,
  M.~N. Stefan J{\o}or~and, A.~Nothhelfer, F.~Petit, J.~Reill, N.~Seitz,
  T.~Wimb{\"o}ck, S.~Wolf, T.~W{\"u}sthoff, and G.~Hirzinger, ``The {DLR} hand
  arm system,'' in \emph{Proceedings of The 2011 IEEE International Conference
  on Robotics and Automation}, 2011, pp. 3175--3182.

\bibitem{icra2014:roboray_hand:kim}
Y.-J. Kim, Y.~Lee, J.~Kim, J.-W. Lee, K.-M. Park, K.-S. Roh, and J.-Y. Choi,
  ``Roboray hand: A highly backdrivable robotic hand with sensorless contact
  force measurements,'' in \emph{2014 IEEE International Conference on Robotics
  and Automation (ICRA)}, May 2014, pp. 6712--6718.

\bibitem{ijrr2016:Deimel:compliant_hand}
R.~Deimel and O.~Brock, ``A novel type of compliant and underactuated robotic
  hand for dexterous grasping,'' \emph{The International Journal of Robotics
  Research}, vol.~35, pp. 161--185, 2016.

\bibitem{icra2017:compliant_hand:Wiste}
T.~Wiste and M.~Goldfarb, ``Design of a simplified compliant anthropomorphic
  robot hand,'' in \emph{2017 IEEE International Conference on Robotics and
  Automation (ICRA)}, May 2017, pp. 3433--3438.

\bibitem{icra2016:ligament_hand:xu}
Z.~Xu and E.~Todorov, ``Design of a highly biomimetic anthropomorphic robotic
  hand towards artificial limb regeneration,'' in \emph{2016 IEEE International
  Conference on Robotics and Automation (ICRA)}, May 2016, pp. 3485--3492.

\bibitem{takaki:high_power_hand}
T.~Takaki and T.~Omata, ``High-performance anthropomorphic robot hand with
  grasping-force-magnification mechanism,'' \emph{IEEE/ASME Transactions on
  Mechatronics}, vol.~16, no.~3, pp. 583--591, June 2011.

\bibitem{humanoids2016:asano:kengoro}
Y.~Asano, T.~Kozuki, S.~Ookubo, M.~Kawamura, S.~Nakashima, T.~Katayama,
  Y.~Iori, H.~Toshinori, K.~Kawaharazuka, S.~Makino, Y.~Kakiuchi, K.~Okada, ,
  and M.~Inaba, ``Humanmimetic musculoskeletal humanoid kengoro toward real
  world physically interactive actions,'' in \emph{Proceedings of the 2016
  IEEE-RAS International Conference on Humanoid Robots}, 2016, pp. 876--883.

\bibitem{HandDesign:Makino:IROS2017}
S.~Makino, K.~Kawaharazuka, M.~Kawamura, Y.~Asano, K.~Okada, and M.~Inaba,
  ``\BIBforeignlanguage{english}{High-power, flexible, robust hand: Development
  of musculoskeletal hand using machined springs and realization of self-weight
  supporting motion with humanoid},'' in
  \emph{\BIBforeignlanguage{english}{Proceedings of The 2017 IEE/RSJ
  International Conference on Robotics and Systems}}, september 2017, pp.
  1187--1192.

\bibitem{ForearmDesign:Kawaharazuka:IROS2017}
K.~Kawaharazuka, S.~Makino, M.~Kawamura, Y.~Asano, Y.~Kakiuchi, K.~Okada, and
  M.~Inaba, ``\BIBforeignlanguage{english}{Human mimetic forearm design with
  radioulnar joint using miniature bone-muscle modules and its applications},''
  in \emph{\BIBforeignlanguage{english}{Proceedings of The 2017 IEE/RSJ
  International Conference on Robotics and Systems}}, september 2017, pp.
  1164--1171.

\bibitem{HandClutteredNarrow:Hasegawa:IROS2017}
S.~Hasegawa, K.~Wada, Y.~Niitani, K.~Okada, and M.~Inaba,
  ``\BIBforeignlanguage{english}{A three-fingered hand with a suction gripping
  system for picking various objects in cluttered narrow space},'' in
  \emph{\BIBforeignlanguage{english}{Proceedings of The 2017 IEE/RSJ
  International Conference on Robotics and Systems}}, september 2017, pp.
  1164--1171.

\bibitem{cutkosky:grasp_choice}
M.~R. Cutkosky, ``Grasp choice, grasp models and the design of hands for
  manufacturing tasks,'' \emph{IEEE Transactions on Robotics and Automation},
  vol.~5, no.~3, pp. 269--279, 1989.

\bibitem{MuscleModule:Asano:IROS2015}
Y.~Asano, T.~Kozuki, S.~Ookubo, K.~Kawasaki, T.~Shirai, K.~Kimura, K.~Okada,
  and M.~Inaba, ``\BIBforeignlanguage{english}{A sensor-driver integrated
  muscle module with high-tension measurability and flexibility for
  tendon-driven robots},'' Oct 2015, pp. 5960--5965.

\end{thebibliography}

\end{document}